\documentclass[sigconf,authorversion]{acmart}
\usepackage{fancyhdr}
\AtBeginDocument{%
    \addtolength{\footskip}{2.0\baselineskip}%
    \fancyfoot[L]{\textit{\textbf{Preprint --- do not distribute.}}}%
}

\usepackage{xcolor}
\usepackage{array}
\usepackage{multirow}
\newcolumntype{C}[1]{>{\centering\let\newline\\\arraybackslash\hspace{0pt}}m{#1}}

\graphicspath{{figure/}{figures/}}

\copyrightyear{2023}
\acmYear{2023}
\setcopyright{none}
\acmConference[]{}{}{}
\acmBooktitle{4th ACM International Conference on AI in Finance (ICAIF '23),
November 27--29, 2023, Brooklyn, NY, USA}
\acmPrice{15.00}
\acmDOI{10.1145/3604237.3626850}
\acmISBN{979-8-4007-0240-2/23/11}

\begin{document}
\title{Towards a Foundation Purchasing Model: Pretrained Generative Autoregression on Transaction Sequences}

\author{Piotr Skalski}
\orcid{0000-0003-3102-9837}
\affiliation{%
  \department{Innovation Lab}
  \institution{Featurespace}
  \city{Cambridge} 
  \country{UK} 
}
\email{piotr.skalski@featurespace.co.uk}

\author{David Sutton}
\orcid{0009-0005-6739-5689}
\affiliation{%
  \department{Innovation Lab}
  \institution{Featurespace}
  \city{Cambridge} 
  \country{UK} 
}
\email{david.sutton@featurespace.co.uk}

\author{Stuart Burrell}
\orcid{0000-0002-6333-1750}
\affiliation{%
  \department{Innovation Lab}
  \institution{Featurespace}
  \city{Cambridge} 
  \country{UK} 
}
\email{stuart.burrell@featurespace.co.uk}

\author{Iker Perez}
\orcid{0000-0001-9400-4229}
\affiliation{%
  \department{Innovation Lab}
  \institution{Featurespace}
  \city{Cambridge}
  \country{UK} 
}
\email{iker.perez@featurespace.co.uk}

\author{Jason Wong}
\orcid{0000-0001-7727-1341}
\affiliation{%
  \department{Innovation Lab}
  \institution{Featurespace}
  \city{Cambridge} 
  \country{UK}
}
\email{jason.wong@featurespace.co.uk}

\renewcommand{\shortauthors}{Skalski, et al.}

\begin{abstract}

Machine learning models underpin many modern financial systems for use cases such as fraud detection and churn prediction. Most are based on supervised learning with hand-engineered features, which relies heavily on the availability of labelled data. Large self-supervised generative models have shown tremendous success in natural language processing and computer vision, yet so far they haven't been adapted to multivariate time series of financial transactions. In this paper, we present a generative pretraining method that can be used to obtain contextualised embeddings of financial transactions. Benchmarks on public datasets demonstrate that it outperforms state-of-the-art self-supervised methods on a range of downstream tasks. We additionally perform large-scale pretraining of an embedding model using a corpus of data from 180 issuing banks containing 5.1 billion transactions and apply it to the card fraud detection problem on hold-out datasets. The embedding model significantly improves value detection rate at high precision thresholds and transfers well to out-of-domain distributions.

\end{abstract}

%
%

\ccsdesc[500]{Applied computing~Online banking}
\ccsdesc[500]{Computing methodologies~Unsupervised learning}
\ccsdesc[500]{Computing methodologies~Learning latent representations}

\keywords{transaction embeddings, self-supervised learning, generative modelling, multivariate time series, fraud detection}

\maketitle

\section{Introduction}

Foundation models have seen tremendous success and wide adoption within the past couple of years. They have proven their ability to leverage large corpora of data and scale to hundreds of billions of parameters. On textual data, these models can be used not only to generate human-level text but also to produce contextualised embeddings of individual tokens, sentences, and even whole documents that can be fed as inputs to downstream models. Their rapid success has been in no small part due to the development of self-supervised learning (SSL) methods such as autoregressive~\cite{gpt} and masked~\cite{bert} language modelling which have allowed models to learn contextual representations of input tokens without relying on labels.

While these methods have already been successfully used with different modalities such as natural language \citeN{gpt,gpt2,gpt3,albert,electra}, computer vision \citeN{selfie,contextencoder}, audio \citeN{wav2vec,wav2vec2}, and tabular data \citeN{tabnet,tabtransformer, subtab} there has been little work to adapt them to the case of multivariate time series data. One example of such data modality of particular interest in this work is streams of financial transactions -- sequences of events representing transfers of funds between two entities. Each event can be described by a set of numerical or categorical features, such as the timestamp, card number, transaction amount, merchant name, or merchant category (in the case of card transactions).

From the perspective of financial institutions, the most important modelling problems in this domain include fraud detection, money laundering detection, credit default prediction, customer churn prediction, and future expenditure modelling. Most common approaches to solve these problems are based on supervised learning and rely on hand-engineered features which take time and domain expertise to define for specific modelling problems. These approaches are therefore not amenable to transfer learning and require redesigning of feature definitions when new fraud typologies emerge.

Self-supervised learning has the potential to replace the expensive feature engineering process in favour of learnt representations from a foundation model pretrained on large quantities of unlabelled data. However, efforts in this space have so far been limited. Recently, a contrastive learning SSL approach was designed to generate embeddings of cardholders based on their transaction history ~\cite{coles}. These embeddings were evaluated on entity-level classification tasks and shown to perform on par and in some cases better than hand-engineered features. However, autoregressive language modelling approaches have not yet been adapted to the domain of financial transactions, even though the task of predicting future events bears a close resemblance to modelling problems in the financial industry.

The success of generative pretraining methods in NLP and lack of equivalent approaches in the domain of financial transactions has motivated our research. In this paper, we present a self-supervised learning method for pretraining autoregressive models that can generate \textit{transaction embeddings}. Our pretraining method \textit{NPPR} combines two objectives: \textit{next event prediction (NP)} and \textit{past reconstruction (PR)}. The next event prediction task, adapted from language modelling to handle multivariate transaction events, was motivated by the similarity between generative modelling and the financial modelling tasks such as churn, credit default and expenditure prediction. All of these tasks aim to predict the future actions by an entity, and solving them requires the model to encode features capturing behavioral characteristics of entities. The past reconstruction task serves the purpose of further encouraging the model to learn longer-term behavioral features which increase the predictive performance of the embeddings on downstream problems.

We evaluate our method on four publicly available datasets of card transactions, showing that the generated embeddings can outperform hand-engineered features and other SSL methods on churn prediction, age group classification, expenditure forecasting, and credit default prediction. We furthermore use our method to pretrain a \textit{Foundation Purchasing Model} on a large corpus of transaction histories from 180 European issuing banks and use the model to produce transaction embeddings on three hold-out issuer datasets that were excluded from the pretraining corpus. The hold-out issuers operate in a different country to any of the pretraining issuers. We apply these embeddings to the fraud detection problem, showing transferability of the model to significantly out-of-domain data and benefits of pretraining on a large and diversified corpus of transactions. Visualisations of the embedding space show that the model encodes similarity among different merchant category codes akin to semantic similarity of word embeddings learnt by large language models.

To summarise, in this paper we make the following contributions:
\begin{enumerate}
    \item propose a self-supervised learning method that combines a next event prediction task with a past reconstruction task, both adapted to the domain of multivariate time series of financial transactions;
    \item show that our method outperforms hand-engineered features and other pretraining methods on downstream classification and regression tasks using evaluations on public datasets;
    \item demonstrate that pretraining with our method on a large corpus of card transaction datasets from 180 issuing banks improves fraud detection at high precision thresholds and transfers well to out-of-domain data;
    \item illustrate that the resulting embeddings are able to capture semantic similarity between merchant category codes.
\end{enumerate}

\section{Related Work}

Many SSL tasks for sequential data were originally designed for the domain of natural language. Autoregressive language modelling aims to predict the next token in a sentence based on the previous ones and has been used to pretrain the GPT family of models \citeN{gpt,gpt2,gpt3}. In masked language modelling (MLM) \cite{bert}, randomly sampled tokens are masked with a special mask token and the network is tasked with predicting the original token. This method has been successfully adapted to other domains including vision \citeN{selfie,contextencoder}, audio \citeN{wav2vec,wav2vec2}, and tabular data \citeN{tabnet,tabtransformer}. Next sentence prediction \cite{bert} has been used together with MLM and works by feeding the network two sentences A and B and predicting whether B follows A. Replaced token detection, used by ELECTRA \cite{electra}, is a modification of MLM where randomly sampled tokens are replaced with candidates generated by a different language model and the task is to predict the original input token.

Another popular class of SSL methods is contrastive learning. Typically, it learns representations that are invariant to data augmentation. It involves generating positive and negative pairs where the positive pairs come from two augmented views of the same sample, while negative pairs come from two different samples. A contrastive loss function encourages representations to be similar for positive pairs and dissimilar for negative pairs. Some examples include SimCLR \cite{simclr} for images (which uses a composition of standard image augmentation methods), SAINT \cite{saint} for tabular data (uses CutMix \cite{cutmix} in input space and mixup \cite{mixup} in latent space), SimCSE \cite{simcse} (applies dropout as data augmentation). CPC \cite{cpc} is a variation of contrastive learning applicable to autoregressive models that tries to maximize the mutual information between the hidden state and future events from the same sequence. In the domain of financial transactions, to the best of our knowledge, CoLES \cite{coles} is the only method that has used contrastive learning to obtain entity embeddings. It uses randomly generated subsequences from a transaction history belonging to the same entity as positive pairs and subsequences from different entity histories as negative pairs.

There also exist non-contrastive methods which train representations invariant to data augmentation using positive examples only. They avoid representation collapse by using a momentum encoder (BYOL \cite{byol}, TiCo \cite{tico}), penalizing cross-correlation between positive views (Barlow Twins \cite{barlowtwins}), clustering embeddings with an equipartition constraint (SwAV \cite{swav}), and applying an asymmetrical stop-gradient operation (SimSiam \cite{simsiam}).

\begin{figure*}
    \includegraphics[width=0.9\textwidth]{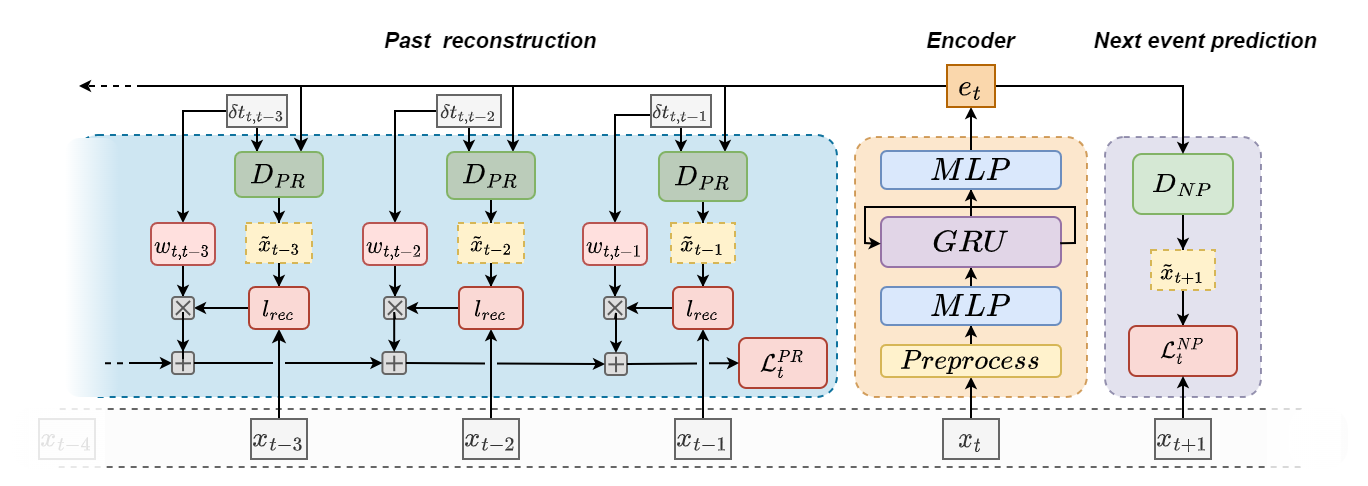}
    \caption{The NPPR generative modelling framework for pretraining a recurrent encoder using a combination of \textit{next event prediction} and \textit{past reconstruction} tasks.}
    \Description{Illustration of the NPPR generative modelling framework for pretraining a recurrent encoder using a combination of next event prediction and past reconstruction tasks.}
    \label{fig:pretraining}
\end{figure*}

Financial transactions have also been used with graph-based methods where originators and beneficiaries (and sometimes transactions) form the nodes of a graph. Some early work in this area involve supervised training for fraud detection \citeN{leuven,G-molloy,G-mastercard} and generating merchant embeddings \citeN{deeptrax,deeptrax2}. More recently, there has also been progress in inductive representation learning. GraphSage \cite{graphsage} encourages similar representations for nearby nodes and has been used to create embeddings for credit card fraud detection \citeN{leuven2,leuven3}. Link prediction between nodes has been used as an SSL task for detecting anomalous transactions \cite{laundrograph} and capturing inter-company relationships \cite{G-minikawa}. However, none of these methods leverages the inherent time-series nature of transactions. They are typically investigated in the context of money laundering detection, where patterns of movement of funds across a network are analysed periodically in a batch process. In contrast, in this work, we are interested in generating transaction embeddings for applications that require real-time processing, such as fraud prevention.

\section{Generative modelling on transaction sequences}

\subsection{Proposed method}

Suppose $\mathcal{H}=\{h_e\}$ is a set of transaction histories pertaining to some entities such as cardholders or account holders. A transaction history is a time-ordered sequence $h_e=\{x_t\}_{t=0}^{T_e}$ of financial transactions, where each transaction $x_t \in \mathcal{X} $ is described by a set of numerical and categorical features (amount, merchant name, etc.).~\footnote{For clarity of notation the subscript $e$ on transactions has been dropped.} The goal is to train an encoder network $E \colon \mathcal{X}^*\to\mathbb{R}^d$ that creates an embedding vector $e_t\in\mathbb{R}^d$ of a transaction $x_t$ given past transactions from the same entity up to this one, i.e. $e_t=E(x_t, x_{t-1}, ..., x_0)$. To extend this setup to any multivariate time series data, we will refer to transaction histories as \textit{sequences} and individual transactions as \textit{events}. The proposed self-supervised algorithm is visualised in Figure~\ref{fig:pretraining} and is composed of two tasks.

\textbf{Next event prediction (NP)} is the primary task and adapts autoregressive language modelling to the case of multivariate events. A decoder network $D_{NP} \colon \mathbb{R}^d\to\mathcal{X}$ takes an embedding $e_t$ of event $x_t$ to generate predictions of the next event's features $\tilde{x}_{t+1} = D_{NP}(e_t)$. For numerical features the predictions are simply real numbers while for categorical features they are vectors of probabilities over the distinct categories of the feature. The objective function is defined as
\begin{equation}
    \mathcal{L}_t^{NP}=\sum_{f} l^f_{rec}\left((\tilde{x}_{t+1})_f, (x_{t+1})_f\right)
\end{equation}
where $()_f$ denotes a slice of a vector corresponding to a particular feature $f$ and the reconstruction loss function $l^f_{rec}$ for a single feature $f$ is mean squared error if $f$ is a numerical feature and cross-entropy if $f$ is a categorical feature.

\textbf{Past reconstruction (PR)} is the secondary task that aims to guide the encoder towards learning behavioral features with long-term dependencies. We define a decoder network $D_{PR}:(\mathbb{R}^d, \mathbb{R})\to\mathcal{X}$ that takes as input an embedding $e_t$ of event $x_t$ and a scalar time difference $\delta t_{t,t-k}$ between this event and an event from the past $x_{t-k}$ and generates a reconstruction of that past event $\tilde{x}_{t-k}=D_{PR}(e_t, \delta t_{t,t-k})$. The objective function is a weighted sum of reconstruction losses of past events:
\begin{equation}
    \mathcal{L}_t^{PR}=\sum_{k=1}^{min(K,t)} \omega_{t,t-k} \sum_{f} l^f_{rec}\left((\tilde{x}_{t-k})_f, (x_{t-k})_f\right)
\end{equation}
where $\omega_{t,t-k}=exp(-\delta t_{t,t-k}/\lambda)$ is a weight function that decays exponentially with the time difference between events at a rate governed by the decay length hyperparameter $\lambda$. The summation over the past events is truncated so that at most $K$ past events contribute to the loss.

For a sequence of events $h_e$, we define the total loss as the sum of event losses that are weighted combinations of the two objective functions defined above:
\begin{equation}
    \mathcal{L}_e=\sum_t^{T_e} (1-\alpha)\mathcal{L}_t^{NP} + \alpha\mathcal{L}_t^{PR}
\end{equation}
where $\alpha \in (0, 1)$ is a hyperparameter.

\subsection{Model architecture}

Although any autoregressive model architecture can be used as an encoder, we decided to use a recurrent model based on GRUs \cite{gru}. Compared to unidirectional transformer models, RNNs are more efficient in production where new events arrive one at a time since they only have to store and process hidden state and a new event rather than a whole sequence of previous events. This consideration is often very important in the financial sector when using a model for real-time decisioning, where there are often stringent requirements on response latency. The encoder architecture is shown in Figure~\ref{fig:pretraining}. An event $x_t$ is first preprocessed into a dense vector. Numerical features are normalized while categorical features are encoded through an embedding layer. Additionally, event timestamps are used to produce a numerical feature that encodes the time gap between events $x_t$ and $x_{t-1}$ belonging to the same entity. The resulting vectors for individual features are concatenated to form a single dense representation of an event. This vector is then passed through a stack of layers $\phi_{proj}\circ \phi_{GRU}\circ \phi_{MLP}$ that enriches it with the representation of past events. The use of MLP before the GRU layer adds expressivity to the encoder and the projection layer $\phi_{proj}$ allows the size of the hidden state to scale independently of the final embedding size.

The decoder models in the two modelling heads are simple MLPs. The output of the last dense layer (with linear activation) in each decoder is split into multiple vectors, one for each feature in the encoded events. The vectors corresponding to numerical features have size one (scalar), while those corresponding to categorical features have size equal to the number of distinct values the feature can take. The vectors for categorical features are additionally passed through a \textit{softmax} activation to produce probability estimates.

\section{Experiments on public datasets}

In this section we evaluate the performance of entity-level embeddings generated with our method when used in classification and regression tasks. We use the embeddings as inputs to downstream models without fine-tuning the pretrained models. This evaluation setup is suitable for testing systems where the embeddings server and downstream modelling setup are decoupled. Code to reproduce experiments in this section is publicly available on GitHub\footnote{https://github.com/Featurespace/foundation-model-paper}. Since publicly available transaction datasets are very limited and their schemas are often incompatible, pretraining and evaluation was performed separately on each dataset. Pretraining on a large corpus of data and out-of-domain evaluation on hold-out datasets is investigated in the next section using private datasets.

\subsection{Datasets}

We use publicly available datasets from various data science competitions comprising debit and credit card financial transactions. These datasets include unlabelled and labelled transactions for four different tasks: age group prediction\footnote{https://ods.ai/competitions/sberbank-sirius-lesson}, churn prediction\footnote{https://boosters.pro/championship/rosbank1/}, future expenditure (expnd.) forecasting\footnote{https://ods.ai/competitions/x5-retailhero-uplift-modeling}, and credit default prediction\footnote{https://boosters.pro/championship/alfabattle2/overview}. Important statistics of each dataset are shown in Table~\ref{tab:datasets}. For the expenditure forecasting task, we split the original 4-month training period into a 3-month training period and a 1-month labelling period. We then construct one labelled example per entity by taking its transactions over the 3-month training period as the input and using its total expenditure over the 1-month labelling period as the label.

Training and test sets for each dataset were generated by using 80\% of entity histories as the training set and the remaining 20\% as the test set. The pretraining set was constructed by concatenating the unlabelled entity histories and the training set.

\subsection{Hyperparameters}

For pretraining, we used the same encoder architecture on each dataset: MLP with 2 hidden layers (512 neurons each) and ReLU activation followed by a GRU with hidden state size 512, followed by a dense projection layer (with sigmoid activation) to the embedding space of size 512. Both the NP and PR decoders in our method are MLPs with 2 hidden layers (512 neurons each) and ReLU activations. For training, we used the Adam optimiser with learning rate $\mathrm{10^{-3}}$ and early stopping.

For pretraining with our method, the decay length $\lambda$ was chosen to be 2 months based on domain expertise. The hyperparameter $\alpha$ controlling the proportion of past reconstruction task in the total loss was tuned on each dataset separately using cross-validation: 0.1 (churn), 0.001 (age), 0.005 (expenditure), and 0.001 (credit default).
For downstream model training, we used MLPs with 3 hidden layers (512 neurons each on churn and age prediction, and 1024 neurons each on expenditure and credit default prediction) and ReLU activations together with dropout and weight decay regularisation. Dropout rate, weight decay, and learning rate were tuned on each baseline separately using the Optuna framework with 5-fold cross-validation.

\begin{table}
    \caption{Characteristics of the four publicly available datasets of financial transactions.}
    \label{tab:datasets}
    \begin{tabular}{C{1.3cm} C{1.2cm} C{1.6cm} C{1.3cm} C{1.3cm}}
        \toprule
         \textbf{Name} & \textbf{labelled cards} & \textbf{unlabelled cards} & \textbf{num. features} & \textbf{cat. features} \\
        \midrule
         Churn $\mathrm{^{(Rosbank)}}$ & 5K & 5K & 1 & 4 \\
         Age $\mathrm{^{(SberBank)}}$ & 30K & 20K & 1 & 1 \\
         Expnd. $\mathrm{^{(X5\:Group)}}$ & 400K & 0 & 2 & 6 \\
         Credit default $\mathrm{^{(Alpha Bank)}}$ & 960K & 510K & 1 & 14 \\
        \bottomrule
    \end{tabular}
\end{table}

\subsection{Baselines}

We compare our approach against four baselines.

\textbf{Hand-engineered features}. We use the same hand-crafted features as in \cite{coles}. For numerical features, we apply aggregate functions (sum, count, mean, min, max, variance) over all transactions in an entity history. For categorical features, we compute the above aggregates of numerical features within groups of transactions grouped by every unique value of each categorical feature.

\textbf{SimCSE}. This uses sequence embeddings pretrained with contrastive learning where dropout was used as the data augmentation strategy for generating positive pairs. The dropout rate was tuned on each dataset.

\textbf{Replaced event detection (RED)}. This is an adaptation of the \textit{replaced token detection} task used in ELECTRA \cite{electra} where randomly sampled events are swapped for random events from other sequences in a batch, and the decoder is tasked with predicting the sampling mask. We found that a sampling probability of 30\% performed best on downstream tasks.

\textbf{CoLES}. A contrastive learning method where positive samples are random subsequences coming from the same sequence, and negative samples are subsequences from two different sequences. This method is sensitive to the choice of minimum and maximum subsequence sampling lengths. We adopted the same values of these hyperparameters as in the original paper~\cite{coles}.

For each of the methods above, we used the embedding of the most recent event as the sequence embedding. For the NPPR method, we used the average of all event embeddings in a sequence as the sequence embedding, which can improve performance on downstream tasks as shown in the next section.

\subsection{Results}

Below we report results from each of the chosen methods using tuned hyperparameters. We report the mean and standard deviation on test sets from multiple training runs on different folds of the training set.

\subsubsection{Comparison with baseline methods}

Table~\ref{tab:main_results} compares NPPR to the different baseline methods. Our method outperforms other methods on all datasets, including hand-engineered features, which turns out to be the strongest baseline (outperforming CoLES on two datasets, RTD on three datasets and SimCSE on all datasets). NPPR offers significant performance improvements on the churn, expenditure, and credit default prediction problems which aim to predict the future behavior of an entity. This confirms our hypothesis that generative modelling is particularly suitable for learning behavioral features that are predictive of future events. When predicting a static attribute of an entity, such as age group prediction, the contrastive CoLES method is competitive, but our method shows superior performance even in this case. This is achieved due to transaction embeddings averaging, which we demonstrate in section \ref{sec:averaging}.

\begin{table}
    \caption{Evaluation of self-supervised embeddings on downstream tasks. Average test set performance and standard deviation values from multiple runs on different training set folds.}
    \label{tab:main_results}
    \begin{tabular}{C{1.1cm} C{1.5cm} C{1.5cm} C{1.5cm} C{1.5cm}}
        \toprule
         \textbf{Method} & \textbf{Churn} AUC$\uparrow$ & \textbf{Age} Accuracy$\uparrow$ & \textbf{Expnd.} \quad MSLE$\downarrow$ & \textbf{Default} AUC$\uparrow$ \\
        \midrule
         FeatEng & 0.798$_{\pm0.004}$ & 0.626$_{\pm0.002}$ & 0.743$_{\pm0.001}$ & 0.768$_{\pm0.001}$ \\
         SimCSE & 0.650$_{\pm0.006}$ & 0.410$_{\pm0.002}$ & 1.140$_{\pm0.001}$ & 0.630$_{\pm0.001}$ \\
         RTD & 0.827$_{\pm0.003}$ & 0.590$_{\pm0.001}$ & 0.747$_{\pm0.001}$ & 0.765$_{\pm0.001}$ \\
         CoLES & 0.813$_{\pm0.003}$ & 0.633$_{\pm0.002}$ & 0.758$_{\pm0.001}$ & 0.765$_{\pm0.001}$ \\
         NPPR & \textbf{0.845}$_{\pm0.003}$ & \textbf{0.642}$_{\pm0.001}$ & \textbf{0.723}$_{\pm0.001}$ & \textbf{0.798}$_{\pm0.001}$ \\
        \bottomrule
    \end{tabular}
\end{table}

\subsubsection{Importance of constituent tasks}

Table~\ref{tab:ablate_task} shows results of ablating the two constituent tasks from our method. In both cases, an entity embedding was constructed by averaging the transaction embeddings from the whole entity history as in the NPPR method. In general, embeddings pretrained with the next event prediction task perform significantly better than those using just past reconstruction task, except for churn prediction. In fact, using only the next event prediction task outperforms the other baselines from Table~\ref{tab:main_results} on three out of four problems, which demonstrates the strength of vanilla generative modelling for learning behavioral features.

Even though the performance gap between the two tasks can be significant, as in the case of age group prediction, using a combination of both tasks outperforms pretraining with either of the two tasks in isolation on all datasets. Adding even a small amount of past reconstruction loss to the total loss has a positive effect on the performance on all downstream problems. This suggests that the past reconstruction task encourages each transaction embedding to encode longer-term behavioral patterns which the next event prediction task doesn't explicitly do.

Interestingly, the past reconstruction task performed better than next event prediction on churn prediction. We hypothesize that reconstructing past events helps embeddings encode information from further back in time, which in turn allows the churn prediction model to more accurately model decline in transaction velocity. Consequently, the best performing value of $\alpha$, which controls the contribution of the past reconstruction task to the total loss, was larger on churn prediction compared to other datasets.

\begin{table}
    \caption{Ablation study comparing NPPR to \textit{next event prediction (NP)} and \textit{past reconstruction (PR)} tasks used in isolation. Average test set performance and standard deviation values from multiple runs on different training set folds.}
    \label{tab:ablate_task}
    \begin{tabular}{C{1.0cm} C{1.5cm} C{1.5cm} C{1.5cm} C{1.5cm}}
        \toprule
         \textbf{Method} & \textbf{Churn} AUC$\uparrow$ & \textbf{Age} Accuracy$\uparrow$ & \textbf{Expnd.} \quad MSLE$\downarrow$ & \textbf{Default} AUC$\uparrow$ \\
        \midrule
         NPPR & \textbf{0.845}$_{\pm0.003}$ & \textbf{0.642}$_{\pm0.001}$ & \textbf{0.723}$_{\pm0.001}$ & \textbf{0.798}$_{\pm0.001}$ \\
         PR & 0.833$_{\pm0.004}$ & 0.542$_{\pm0.002}$ & 0.747$_{\pm0.001}$ & 0.744$_{\pm0.001}$ \\
         NP & 0.814$_{\pm0.002}$ & 0.630$_{\pm0.002}$ & 0.733$_{\pm0.001}$ & 0.795$_{\pm0.001}$ \\
        \bottomrule
    \end{tabular}
\end{table}

\begin{table}
    \caption{Relative performance difference on the test set between using averaged transaction embedding vs. embedding of a most recent transaction.}
    \label{tab:averaging}
    \begin{tabular}{C{1.0cm} C{1.2cm} C{1.5cm} C{1.5cm} C{1.3cm}}
        \toprule
         \textbf{Method} & \textbf{Churn} AUC$\uparrow$ & \textbf{Age} Accuracy$\uparrow$ & \textbf{Expnd.} \quad MSLE$\downarrow$ & \textbf{Default} AUC$\uparrow$ \\
        \midrule
         NPPR $\mathrm{^{avg\:vs.\:last}}$ & \textcolor{orange}{-0.5\%} & \textcolor{cyan}{+6.1\%} & \textcolor{cyan}{-0.3\%} & \textcolor{cyan}{+0.6\%} \\
         CoLES $\mathrm{^{avg\:vs.\:last}}$ & \textcolor{orange}{-1.0\%} & \textcolor{cyan}{+0.3\%} & \textcolor{orange}{0.0\%} & \textcolor{orange}{-1.8\%} \\
        \bottomrule
    \end{tabular}
\end{table}

\begin{figure*}
    \includegraphics[width=0.33\textwidth]{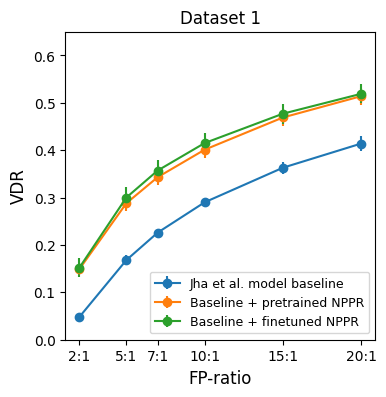}
    \includegraphics[width=0.33\textwidth]{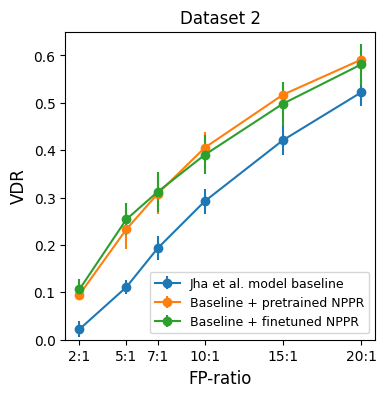}
    \includegraphics[width=0.33\textwidth]{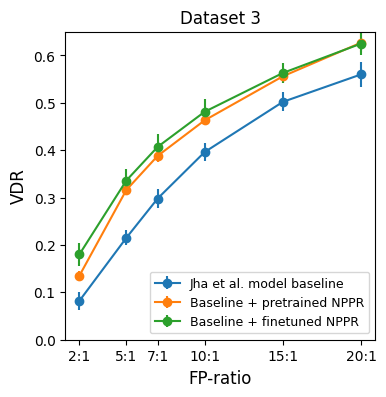}
    \caption{Evaluations of fraud detection models trained on datasets from three different issuers. Three models are shown: baseline hand-engineered features from Jha \textit{et al.}~\cite{fraud-model}, baseline features with NPPR embeddings trained on the pretraining corpus, and baseline features with NPPR embeddings finetuned on downstream datasets. Error bars were computed from multiple independent training runs.}
    \label{fig:fraud}
\end{figure*}

\subsubsection{Effect of averaging transaction embeddings\label{sec:averaging}}

In this experiment, we evaluated the importance of using the average transaction embedding as an entity embedding by comparing it to the strategy adopted in the baseline methods, where the embedding of the most recent transaction was used instead. Table~\ref{tab:averaging} shows the results of this evaluation.

We can see that embedding averaging can improve the performance of our method on downstream problems, especially in cases where the task involves predicting a static entity attribute such as age. However, it can also have a detrimental effect in problems such as churn prediction, presumably because averaging can over-smooth the features encoded in the more recent embeddings which capture a decline in the rate of transacting. By contrast, embedding averaging does not improve CoLES embeddings, which are designed to be similar across transactions by the same entity.

\section{Application to Fraud Detection at scale}

In this section, we apply our self-supervised NPPR method to pretrain a \textit{Foundation Purchasing Model} on transaction data from a large number of issuing banks. We use it to produce transaction embeddings for unseen data, specifically transactions from hold-out issuing banks that operate in different countries to the issuers in the pretraining dataset. A separate fraud classifier is then trained on each of the hold-out issuers. We demonstrate that the pretrained model improves fraud detection performance, transfers well to significantly out-of-domain data, and learns semantic similarity between different merchant categories.

\subsection{Pretrained embedding model}

A single embedding model was pretrained on a corpus of card transaction datasets from 180 European issuing banks, each of which conforms to the ISO 8583 messaging format~\cite{iso}. The corpus contains over 5.1 billion transactions which provide complete transaction histories covering a period of 12 months for 61 million cardholders.

The architecture of the embedding model is as described in the previous sections. It consists of an MLP with 2 hidden layers (2048 neurons each) with ReLU activations, followed by a GRU layer with state size 1024 and a final projection layer to an embedding space of size 768. The decay length $\lambda$ in the NPPR method was 2 months, and the weight of past reconstruction task $\alpha$ was 0.001.

\subsection{Fraud detection models}

\subsubsection{Datasets}

For the downstream fraud detection task, we used labelled datasets from three European issuing banks. These datasets were not part of the pretraining corpus and correspond to issuers that operate in different countries to any of the issuers from the pretraining corpus. This presents the opportunity for testing transfer of the embedding model to significantly out-of-domain data.

The three datasets contain 11 months of transactions from 17 million, 3.5 million, and 1.8 million cardholders. We split each dataset into training, validation, and test sets both temporally and on the entity level, i.e. they contain transactions from different cardholders and non-overlapping consecutive time periods.

The fraud rate in each dataset is respectively 0.04\%, 0.027\%, and 0.11\%. Due to the high class imbalance, we downsampled genuine transactions before training classification models (but no downsampling was performed for pretraining).

\subsubsection{Baseline features}

As a baseline, we used a representative traditional fraud prevention model drawn from literature \cite{fraud-model}. It consists of primary transaction attributes and 14 hand-engineered behavioral features in the form of aggregations over windows of past transactions by the same entity at different time scales. Examples of such features include the average amount spent per transaction over the last month, or the total number of transactions with the same merchant during last month. We refer readers to the source paper for a detailed description of the features.

\subsubsection{Models}

Downstream classification models are MLPs with 3 hidden layers (1024 neurons each) and ReLU activations. They were trained with learning rate $10^{-3}$, batch size 1024 and early stopping. Dropout was used as regularisation with rate 0.2.

\subsubsection{Evaluation metrics}

Production fraud prevention systems are often judged by their performance in reducing fraud losses while operating at high precision score thresholds. False positive predictions lead to declined transactions, which cause losses to the issuing bank and have a detrimental effect on consumer experience. An appropriate metric should measure the value of fraudulent transactions that have been prevented at a certain threshold of declined genuine transactions. A typical metric is the \textit{VDR @ FP-ratio} where \textit{value detection rate (VDR)} is the true positive rate weighed by transaction value and the threshold metric \textit{false positive ratio (FP-ratio)} is the number of false positives divided by the number of true positives. The classification score threshold is adjusted to reach a typical value of FP-ratio between 1:1 and 20:1.

\begin{figure}
    \includegraphics[width=0.48\textwidth]{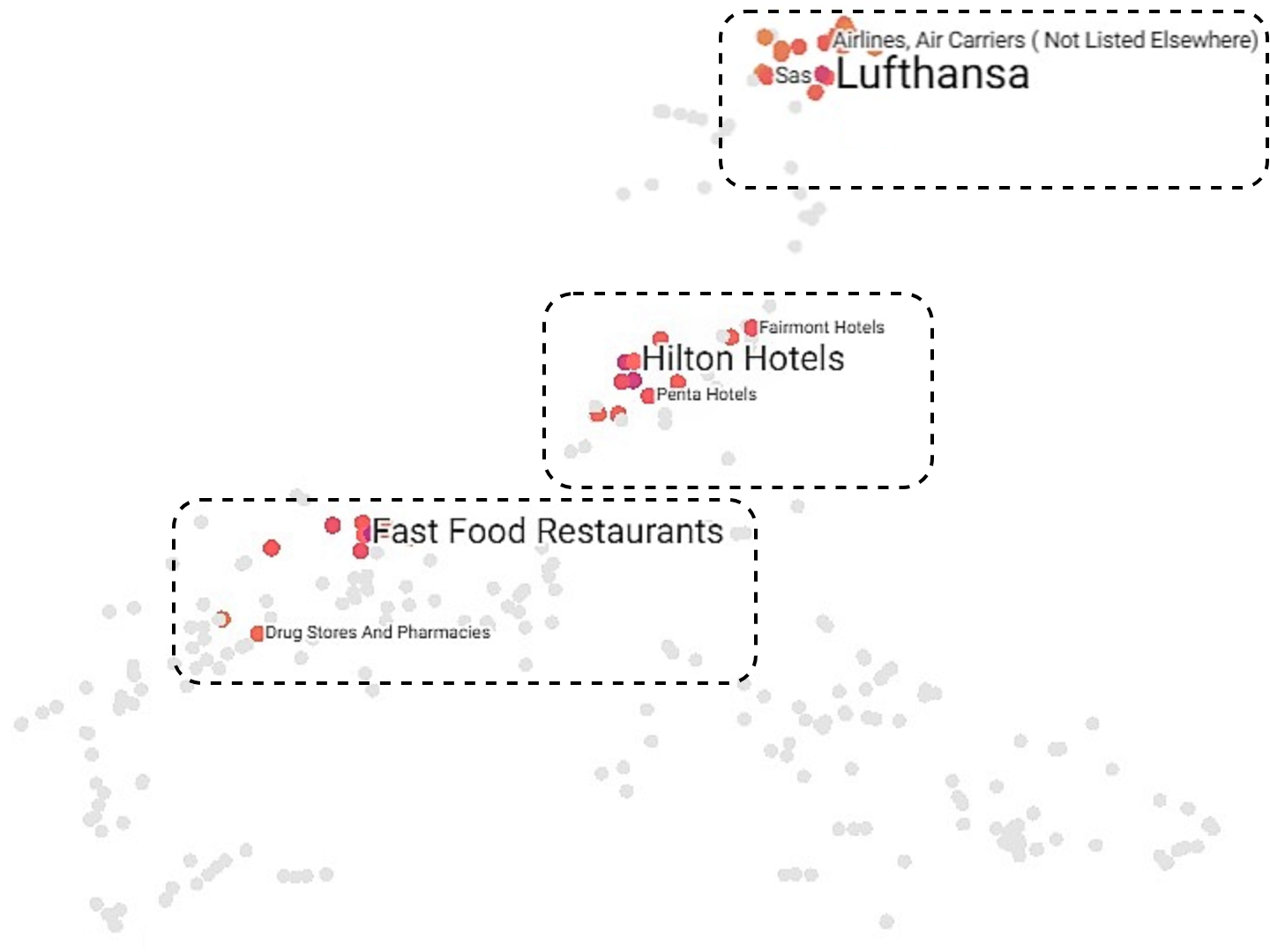}
    \caption{t-SNE projection of a MCC embedding space. Each MCC embedding was obtained by averaging transaction embeddings corresponding to those MCCs.}
    \label{fig:mcc-space}
\end{figure}

\subsection{Quantitative results}

We evaluated three different classification models comparing baseline features, baseline features with NPPR embeddings from the pretrained model, and baseline features with NPPR embeddings from the pretrained model that has been finetuned on downstream datasets. Finetuning was performed in a self-supervised way using our NPPR method and is therefore applicable to the scenario where embeddings server and downstream modelling setup are decoupled. Results are shown in Figure~\ref{fig:fraud} where VDR is plotted against different FP-ratio thresholds.

The addition of NPPR embeddings to the baseline features provides significant improvements in the value detection rate on all FP-ratio thresholds. At 5:1 FP-ratio our embeddings can provide up to 140\% uplift over the hand-engineered features. On all hold-out datasets, embeddings generated by the pretrained model show comparable performance to embeddings generated by finetuned models. This demonstrates the effectiveness of pretraining on a large corpus of diverse datasets and transferability of the pretrained model to significantly out-of-domain data.

\subsection{Visualising the embedding space}

To provide insights into the information encoded by embeddings, we provide visualisations of embeddings of merchant category code (MCC). These codes classify merchants and businesses by the type of goods or services provided. Large merchants classified as airlines, car rental companies and lodging providers typically have their own MCC. Since the input data does not provide any extra information relating these codes to each other, any emergent structure in the embedding space comes entirely from similarities in purchasing behaviors learnt by the model. Each MCC embedding was calculated as the average of all transaction embeddings corresponding to that MCC. Figure~\ref{fig:mcc-space} shows a t-SNE projection of the MCC embedding space together with three selected MCCs and their nearest neighbours. Table~\ref{tab:mcc_neighbours} lists the five nearest neighbours and their distances (measured by cosine distance) to each selected MCC.

We can see that the nearest neighbours of Lufthansa are all airline companies, while those of Hilton Hotels belong to the lodging industry, i.e. embeddings of merchants in the same industry are located close to each other in the embedding space. The cluster of embeddings for hotels is in close proximity to the cluster for airlines, which is expected since purchases from merchants in these two clusters are often correlated. This illustrates that generative modelling allows the embedding space to encode meaningful similarity between different MCCs akin to semantic similarity captured by word embeddings from large language models. This observation is consistent with \cite{deeptrax}, where merchant node embeddings from a graph neural network were used.

In a similar fashion, aggregations of transaction embeddings could be used to obtain embeddings of other entities, such as merchants, cardholders, transaction types, and geographical locations. They can potentially be used as features to support decision-making in recommendation engines and other financial systems.

\begin{table}
    \caption{Nearest neighbours in the embedding space of three MCC embeddings. For each nearest neighbour MCC we show cosine distance in the original space between the MCC in the top row.}
    \label{tab:mcc_neighbours}
    \begin{tabular}{C{0.1cm} C{2.3cm} C{2.3cm} C{2.3cm}}
        \toprule
          & \textbf{Lufthansa} & \textbf{Hilton Hotels} & \textbf{Fast Food} \\
        \midrule
         \multirow{8}{1.7cm}{\rotatebox[origin=c]{90}{\textbf{Nearest neighbours}}} & British Airways $\mathrm{^{(0.23)}}$ 
          & Doubletree Hotels $\mathrm{^{(0.19)}}$ & Eating places $\mathrm{^{(0.14)}}$ \\
           & Scandinavian Airlines\quad\quad $\mathrm{^{(0.32)}}$ & Hampton Inns $\mathrm{^{(0.20)}}$ & Convenience stores\quad\quad\quad $\mathrm{^{(0.26)}}$ \\
           & Air France\quad $\mathrm{^{(0.33)}}$ & Fairmount Hotels $\mathrm{^{(0.28)}}$ & Bakeries\quad\quad $\mathrm{^{(0.27)}}$ \\
           & Swissair\quad\quad $\mathrm{^{(0.35)}}$ & Penta Hotels $\mathrm{^{(0.29)}}$ & News Dealers $\mathrm{^{(0.30)}}$ \\
           & Turkish Airlines $\mathrm{^{(0.38)}}$ & Marriott Hotels $\mathrm{^{(0.32)}}$ & Drug Stores\quad $\mathrm{^{(0.35)}}$ \\
        \bottomrule
    \end{tabular}
\end{table}

\section{Conclusions}

In this paper, we present a self-supervised generative method for obtaining contextualised embeddings of financial transactions by combining two pretraining tasks: next event prediction and past reconstruction. Evaluations on publicly available datasets show that embeddings produced with this method outperform embeddings from other self-supervised methods and hand-engineered features on a range of downstream tasks. We apply our method to the card fraud detection problem and show that it significantly improves the value detection rate at high-precision thresholds. By pretraining on a large corpus of data from multiple issuing banks, we demonstrate that pretrained models trained with our method generalise well to significantly out-of-distribution data.

Pretraining generative models on large textual datasets has led to a class of Foundation Models that abstract away the complexity of natural language modelling in modern AI applications. Likewise, our results on transaction sequences indicate that generative modelling encodes human purchasing behavior in a way that transfers effectively to diverse tasks and out-of-domain data. These properties may enable financial modelling applications to homogenize around a common component - a Foundation Model - which is trained on a large corpus of unlabelled data and which abstracts away the complexity of modelling financial behaviors. This motivates further research on questions of privacy, bias and the potential for few-shot learning, which we defer to future work. 

\begin{acks}
We thank Featurespace for the financial support and resources offered during the completion of this research. We also thank the NVIDIA Inception Program for generous support related to GPU hardware.
\end{acks}

\bibliographystyle{ACM-Reference-Format}
\bibliography{sigproc} 

\end{document}